\def\year{2019}\relax
\documentclass[letterpaper]{article} %DO NOT CHANGE THIS
\usepackage{aaai19}  %Required
\usepackage{times}  %Required
\usepackage{helvet}  %Required
\usepackage{courier}  %Required
\usepackage{url}  %Required
\usepackage{graphicx}  %Required
\usepackage{amsmath}
\usepackage{multirow}
\usepackage{xspace}
\makeatletter
\DeclareRobustCommand\onedot{\futurelet\@let@token\@onedot}
\def\@onedot{\ifx\@let@token.\else.\null\fi\xspace}

\def\eg{\emph{e.g}\onedot} 
\def\ie{\emph{i.e}\onedot} 
 
\def\etc{\emph{etc}\onedot} 
\def\wrt{w.r.t\onedot} 

\makeatother

\title{Show, Attend and Read: \\ A Simple and Strong Baseline for Irregular Text Recognition}
\author{
	Hui Li \thanks{The first two authors equally contributed to this work. C. Shen is the corresponding author.
	 }\textsuperscript{\rm 1}
	~ ~ ~ Peng Wang$^{\ast}$\textsuperscript{\rm 2}
	~ ~ ~ Chunhua Shen\textsuperscript{\rm 1}
	~ ~ ~ Guyu Zhang\textsuperscript{\rm 2} \\
	\textsuperscript{\rm 1}Australian Centre for Robotic Vision, The University of Adelaide, Australia%, SA 5000
	\\
	\textsuperscript{\rm 2}School of Computer Science,  Northwestern Polytechnical University, China \\
	E-mail: $\tt hui.li02@adelaide.edu.au$
}

\begin{document}
\maketitle
\begin{abstract}

Recognizing irregular text in natural scene images is challenging due to the large variance in text appearance, such as curvature, orientation and distortion.
Most existing approaches rely heavily on sophisticated model designs and/or extra fine-grained annotations, which, to some extent, increase the difficulty in algorithm implementation and data collection.
In this work, we propose an easy-to-implement strong baseline for irregular scene text recognition, using off-the-shelf neural network components and only word-level annotations. It is composed of a $31$-layer ResNet, an LSTM-based encoder-decoder framework and a 2-dimensional attention module.
Despite its simplicity, the proposed method is robust. It achieves state-of-the-art performance on irregular text recognition benchmarks and comparable results on regular text datasets.
%The code will be released.
 Code is available at:

	{$\tt https://tinyurl.com/ShowAttendRead$}
\end{abstract}

\begin{figure}[t!]
	\begin{center}
		\includegraphics[width=0.91\columnwidth]{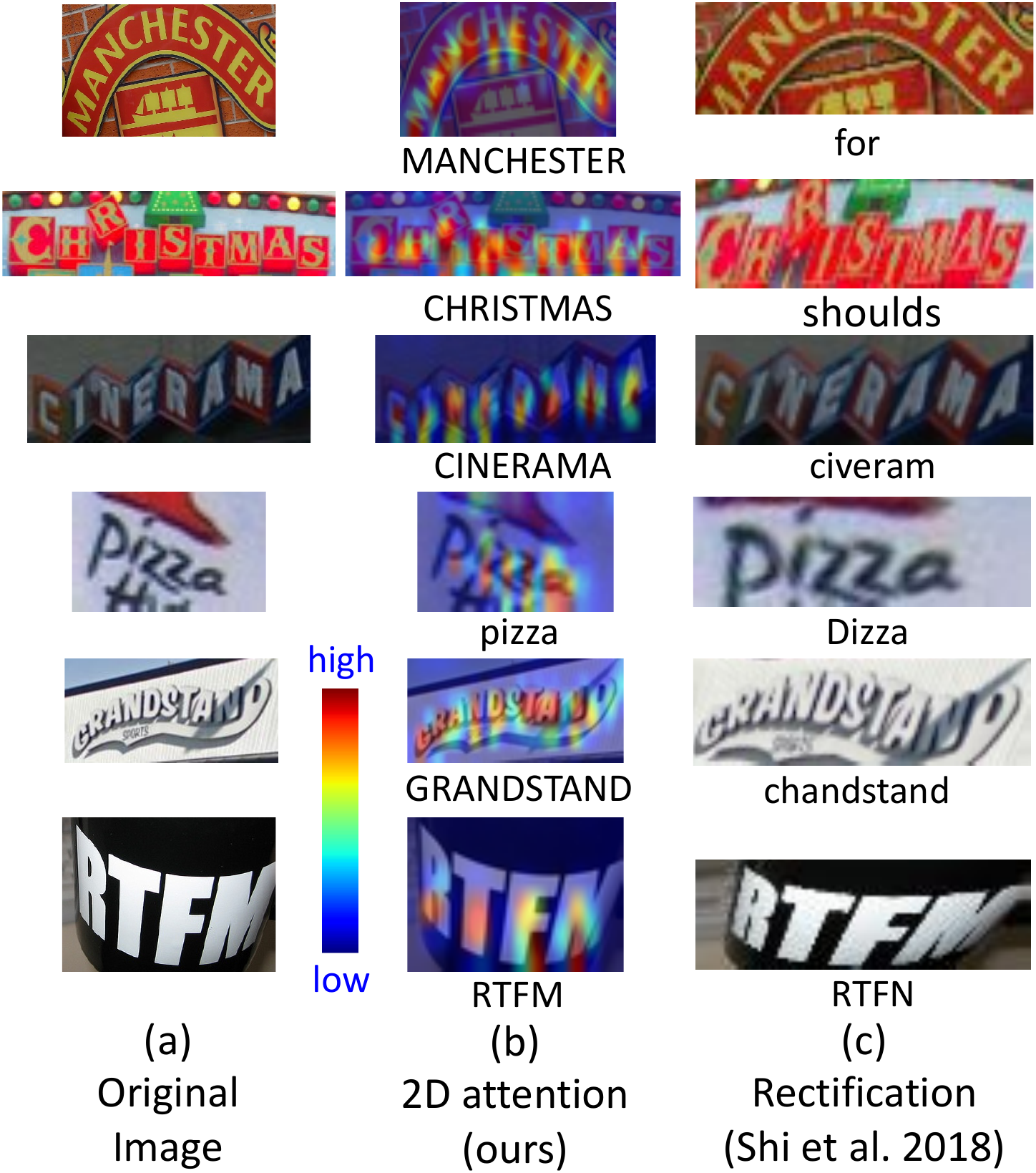}
	\end{center}
	\caption{ The comparison of our proposed $2$D attention based and the rectification based~\cite{shiPAMI2018} irregular text recognizers.
	    The second column gives the predictions of our approach and
		the heat map by aggregating attention weights at all character decoding steps;
		the third column demonstrates the rectified images and the corresponding predictions using the authors' implementation.
		Rectification based methods may encounter difficulties when the input image is severely curved or distorted.
		In contrast, we do not transform images and propose a tailored $2$D attention module to localize individual characters
		in a weakly-supervised manner.
	}
	\label{fig:intro}
\end{figure}

\section{Introduction}
\label{sec:intro}

Text information in images is of indispensable value in semantic visual understanding.
Reading text in natural scene, however, compared to traditional OCR, is still a challenging problem.
One of the main reasons is the potential irregularity and diversity of in text shape and layout, which can be curved, oriented or distorted.
With the application of deep neural networks, the performance of regular (mostly horizontal) text recognition has been improved rapidly. Taking the ICDAR 2013 benchmark~\cite{icdar2013} as example, the best-reported accuracy~\cite{cheng_EditDistance} has been $94.4\%$, to our knowledge.
Nonetheless, most regular text recognizers~\cite{ShiBY15,OCRNIPS17} treat text as horizontal lines, which makes them difficult to be extended directly to irregular text.
The performance of existing irregular text recognizers is far from being satisfactory.
For instance, the current top-performing approach~\cite{shiPAMI2018} only achieves $76.1\%$ accuracy on the ICDAR 2015 benchmark~\cite{icdar2015}.

Existing irregular text recognizers can be roughly categorized into three groups:
rectification based~\cite{shiCVPR2016,shiPAMI2018,BMVC2016_43,Liu2018CharNetAC}, attention based~\cite{ijcai2017,Cheng2017} and multi-direction encoding based~\cite{Cheng2018AON} approaches.
The rectification based methods attempt to transform irregular text patches into regular ones and then recognize them using regular text recognizers.
However, as shown in Figure~\ref{fig:intro}, severe distortions or curvatures give rise to difficulties for rectification.
\cite{ijcai2017} proposed an attention mechanism to select local $2$D features when decoding individual characters.
Nevertheless, it needs extra character-level annotations to supervise the attention network and a multi-task strategy to learn better visual features.
\cite{Cheng2018AON} stated that both rectification and attention based approaches are somewhat difficult to be directly trained on irregular text.
They designed a sophisticated framework that needs to encode arbitrarily-oriented text in four directions.

Alternatively, we go back to the conventional attention based encoder-decoder framework.
Our proposed model is composed of a 31-layer ResNet, an LSTM-based encoder-decoder framework and a tailored $2$-dimensional attention module.
In contrast to \cite{ijcai2017}, our model only needs word-level annotations, which enables us to make full use of both real and synthetic data for training without using character-level annotations.
Built upon standard NN modules, the main architecture can be implemented by around $100$ lines of code.
Despite its simplicity, our method outperforms previous methods on irregular text datasets by a large margin,
and achieves comparable results on regular text. To our best knowledge, we are the first one that uses $2$D attention in irregular text recognition with only word-level annotations needed.
As demonstrated in Figure~\ref{fig:intro}, our $2$D attention module is more flexible and robust in handling sophisticated text layout.

For regular text,
it is presented in \cite{Lee_2016_CVPR,shiCVPR2016} that the $1$D attention based encoder-decoder framework is able to align between input subsequences and decoded characters.
Our approach extends this framework by replacing $1$D attention with a tailored $2$D attention mechanism,
in order to handle the complicated spatial layout of irregular text.
Inspired by the success of the Show-Attend-and-Tell model~\cite{xu2015show} on image captioning,
our model is also based on a $2$D attention based encoder-decoder structure, which is referred to as Show-Attend-and-Read (SAR).
Note that \cite{xu2015show} is designed for image caption, while ours is used for text recognition.

The main contributions of this work is three-fold:

1)	We setup an easy-to-implement strong baseline for recognizing irregular text in natural scene images,
which is made up of off-the-shelf neural components such as CNNs, LSTMs and attention mechanisms.
The proposed model can be trained end-to-end without pre-training.
All the training examples are synthetic or from public real data.
We will release the code and data used for training.

2) Compared to existing irregular text recognizers, our proposed approach does not rely on sophisticated designs (including spatial transformation,
hierarchical attention or multi-directional encoding) to handle text distortions.
Alternatively, we simply use a $2$D attention mechanism to deal with irregular text, which selects local features for individual characters.
Moreover, our proposed attention module does not require additional pixel-level or character-level supervision information,
which is weakly supervised by the cross-entropy loss on the final predictions.
The attention mechanism is also tailored to consider neighborhood information and boosts the recognition performance.

3) Note that many irregular text recognizers perform relatively worse on regular text. In contrast, due to its flexibility and robustness,
the proposed approach not only significantly outperforms existing approaches on irregular text, but also achieves favorable performance on regular text.

\begin{figure*}[t]
	\begin{center}
		\includegraphics[width=0.88872\textwidth]{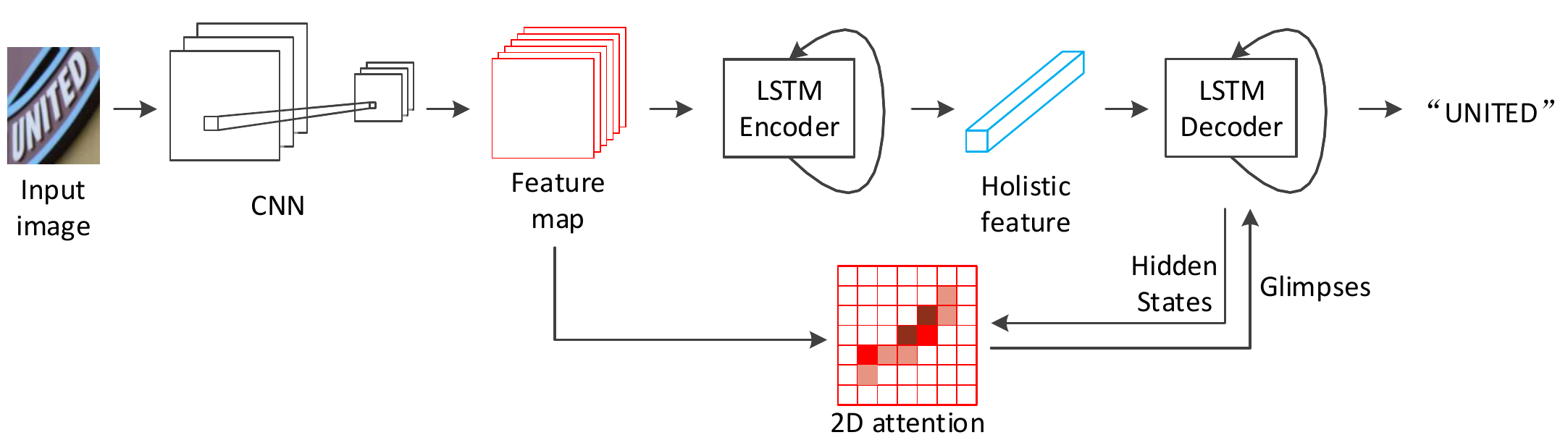}
	\end{center}
	\caption{ Overview of the proposed framework for irregular text recognition. The input image  is firstly fed into a $31$-layer
		ResNet, which results in a $2$D feature map. Next, an LSTM model encodes the feature map column by column, and the last hidden state is considered as a holistic feature of the input image. Another LSTM model is used to decode the holistic feature into a sequence of characters. At each time step of decoding, an attention module computes a weighted sum of $2$D features (glimpse), depending on the current hidden state of the LSTM decoder. The irregularity of text is implicitly handled by the $2$D attention module, in a weakly supervised manner.
	}
	\label{fig:framework}
\end{figure*}

\section{Related Work}
\label{sec:ReWork}
\noindent{\bf Early Work}
Scene text recognition has drawn lots of attentions during recent years and made significant progress in performance.
Early approaches mainly work in a \textit{bottom-up} fashion~\cite{Wangkai2011,MishraCVPR2012,SVTP,YaoCVPR2014}, in which individual characters are detected firstly via sliding window or connected components, and then integrated into a word by dynamic programming or graph models. Character detection or separation by itself, however, is not a completely-solved problem due to complicated background or cursive fonts.
Alternatively, \cite{Max2016IJCV} considered text recognition as a multi-class classification problem, which assigned a distinct label to each word in a $90$k-sized dictionary. Apparently, it is difficult to extend the approach to words out of the dictionary.

\noindent{\bf Regular Text Recognition}
\cite{He2015Reading} and \cite{ShiBY15} considered words as one-dimensional sequences of varying lengths,
and employed RNNs to model the sequences without explicit character separation.
A Connectionist Temporal Classification (CTC) layer was adopted to decode the sequences.
\cite{OCRNIPS17} proposed a Gated Recurrent Convolutional Neural Network (GRCNN) with CTC for regular text recognition.
Inspired by the sequence-to-sequence framework for machine translation, \cite{Lee_2016_CVPR} and \cite{shiCVPR2016} proposed to recognize text using an attention-based encoder-decoder framework. In this manner, RNNs are able to learn the character-level language model hidden in the word strings from the training data. A $1$D soft-attention model was adopted to select relevant local features during decoding characters.
The RNN+CTC and sequence-to-sequence frameworks serve as two meta-algorithms that are widely used by subsequent text recognition approaches.
Both models can be trained end-to-end and achieve considerable improvements on regular text recognition.
\cite{cheng_EditDistance} observed that the frame-wise maximal likelihood loss, which is conventionally used to train the encoder-decoder framework, may be
confused and misled by missing or superfluity of characters, and thus degrade the recognition accuracy.
To this end, they proposed ``Edit Probability'' to handle this misalignment problem.
\cite{SqueezeText18} presented a binary convolutional encoder-decoder network (B-CEDNet) together with a bidirectional recurrent neural network (Bi-RNN) for recognizing regular text images, and achieved significant speed-up.
The whole framework needs to be trained in two stage, and requires pixel-level annotations.
\cite{li2017towards} combined a Faster-RCNN based text detector and a $1$D attention based recognizer into an end-to-end trainable system.

\noindent{\bf Irregular Text Recognition}
The rapid progress on regular text recognition has given rise to increasing attention on recognizing irregular ones.
\cite{shiPAMI2018} and \cite{shiCVPR2016} rectified oriented or curved text based on Spatial Transformer Network (STN)~\cite{jaderberg2015spatial} and
then recognized it using a $1$D attentional sequence-to-sequence model.
\cite{BMVC2016_43} also removed text distortions via STN, and used the RNN + CTC framework for sequence recognition.
Instead of rectifying the entire distorted text image as in \cite{shiPAMI2018,BMVC2016_43}, \cite{Liu2018CharNetAC} presented a Character-Aware Neural Network (Char-Net) to detect and rectify individual characters, which, however, requires extra character-level annotations. Moreover, a sophisticated hierarchical attention mechanism was designed for accurate feature extraction, which consists of a recurrent RoIWarp layer and a character-level attention layer.
\cite{ijcai2017} introduced an auxiliary dense character detection task into the encoder-decoder network to handle the irregular text. Pixel-level character/non-character annotations are required to train the network.
\cite{Cheng2017} asserted that there are ``attention drifts'' in traditional attention model and proposed a Focusing Attention Network (FAN) that is composed of an attention network for character recognition and a focusing network to adjust the attention drift. This work also needs to be trained with character-level bounding box annotations.
\cite{Cheng2018AON} applied LSTMs in four directions to encode arbitrarily-oriented text. A filtering mechanism was designed to integrate these redundant features and reduce irrelevant ones.

\section{Model}
\label{sec:Model}
We describe the architecture of our model in this section. As presented in Figure~\ref{fig:framework}, the whole model consists of two main parts: a ResNet CNN for feature extraction and a $2$D-attention based encoder-decoder model. It takes an image as input and outputs a varying length sequence of characters.

\subsection{ResNet CNN}
The designed $31$-layer ResNet~\cite{ResidualNetwork} is presented in Table~\ref{Tab_CNN}.
For each residual block, we use the projection shortcut (done by $1 \times 1$ convolutions) if the input and output  dimensions are different, and use the identity shortcut if they have the same dimension. All the convolutional kernel size is $3 \times 3$. Besides two $2 \times 2$ max-pooling layers, we also use a $1 \times 2$ max-pooling layer as in~\cite{ShiBY15}, which reserves more information along the horizontal axis and benefits the recognition of narrow shaped characters (\eg, `i', `l'). The resulting $2$D feature maps (denoted as $\mathbf{V}$ of size $H \times W \times D$ where $D$ is the number of channels) will be used: 1) to extract holistic feature for the whole image; 2) as the context for the $2$D attention network.
To keep their original aspect ratios, we resize input images to a fixed height and a varying width.
Hence, the width of the obtained feature map, $W$, also varies \wrt aspect ratios.

\begin{table}[t!]
	\centering
	\resizebox{.65\columnwidth}{!}{
	\begin{tabular}{c|c}
		\hline
		Layer name & Configuration \\
		\hline
		Conv & $3\times 3$, $64$ \\
		\hline
		Conv & $3\times 3$, $128$ \\
		\hline
		Max-pooling & k:$2\times 2$, s:$2\times 2$  \\
		\hline
		Residual block &
			$\left[
			\begin{array}{ccc}
			Conv: 3\times 3,  256 \\
			Conv: 3\times 3,  256 \\
			\end{array}
			\right] \times 1$  \\
		\hline
		Conv& $3\times 3$,  $256$ \\
		\hline
		Max-pooling& k:$2\times 2$, s:$2\times 2$  \\
		\hline
		Residual block &
			$\left[
			\begin{array}{ccc}
			Conv: 3\times 3, 256 \\
			Conv: 3\times 3, 256 \\
			\end{array}
			\right]\times 2$   \\
		\hline
		Conv& $3\times 3$, $256$ \\
		\hline
		Max-pooling& k:$1\times 2$, s:$1\times 2$  \\
		\hline
		Residual block &
			$\left[
			\begin{array}{ccc}
			Conv: 3\times 3, 512 \\
			Conv: 3\times 3, 512 \\
			\end{array}
			\right]\times 5$ 		\\

		\hline
		Conv& $3\times 3$, $512$ \\
		\hline
		Residual  block &
			$\left[
			\begin{array}{ccc}
			Conv: 3\times 3, 512 \\
			Conv: 3\times 3, 512 \\
			\end{array}
			\right]\times 3$  \\
		\hline
		Conv& $3\times 3$, $512$ \\
		\hline
	\end{tabular}
					}
	\caption{The configuration of the $31$-layer ResNet for feature extraction. ``Conv'' stands for Convolutional layers, with kernel size and output channels presented. The stride and padding for convolutional layers are all set to ``1''. For Max-pooling layers, ``k'' means kernel size, and ``s'' represents stride. No padding for Max-pooling layers.}
	\label{Tab_CNN}

\end{table}

\subsection{$2$D Attention based Encoder-Decoder}

Sequence-to-sequence models have been widely used in machine translation, speech recognition and text recognition~\cite{DBLP:conf/nips/Sutskever14}~\cite{Attention15}~\cite{Cheng2017}.
In this work, we adopt a $2$D attention based encoder-decoder network for irregular text recognition.
Without transforming original text images, the proposed attention module is able to accommodate text of arbitrary shape, layout and orientation.

\begin{figure}[t]
	\begin{center}
		\includegraphics[width=0.85\columnwidth]{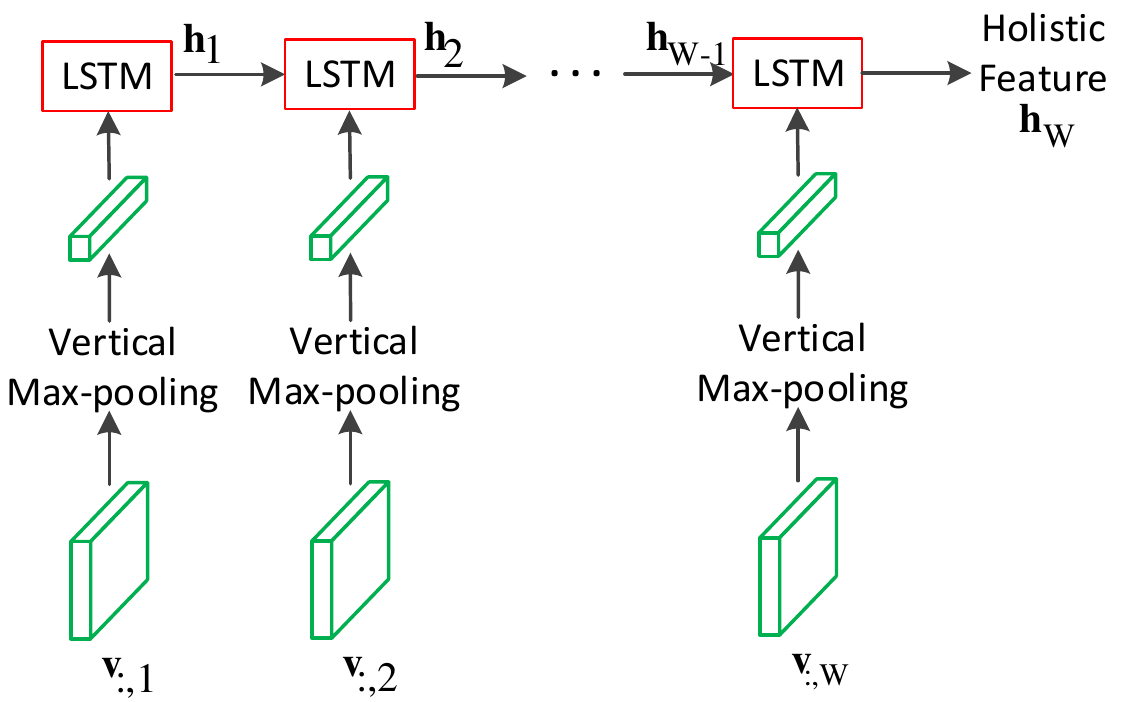}
	\end{center}
	\caption{The structure of the LSTM encoder used in this work. $\mathbf{v}_{:,i}$ represents the $i$th column of the $2$D feature map $\mathbf{V}$.
		At each time step, a column feature is firstly max pooled along the vertical direction, and then fed into LSTM.
	}
	\label{fig:encoder}

\end{figure}

\noindent{\bf Encoder}
As shown in Figure~\ref{fig:encoder},
the encoder is a $2$-layer LSTM model with $512$ hidden state size per layer.
At each time step, the LSTM encoder receives one column of the $2$D features maps followed by max-pooling along the vertical axis,
and updates its hidden state $\mathbf{h}_t$.
After $W$ steps, the final hidden state of the second LSTM layer, $\mathbf{h}_{W}$, is regarded as a fixed-size representation (holistic feature) of the input image,
and provided for decoding.

\begin{figure}[t]
	\begin{center}
		\includegraphics[width=0.88\columnwidth]{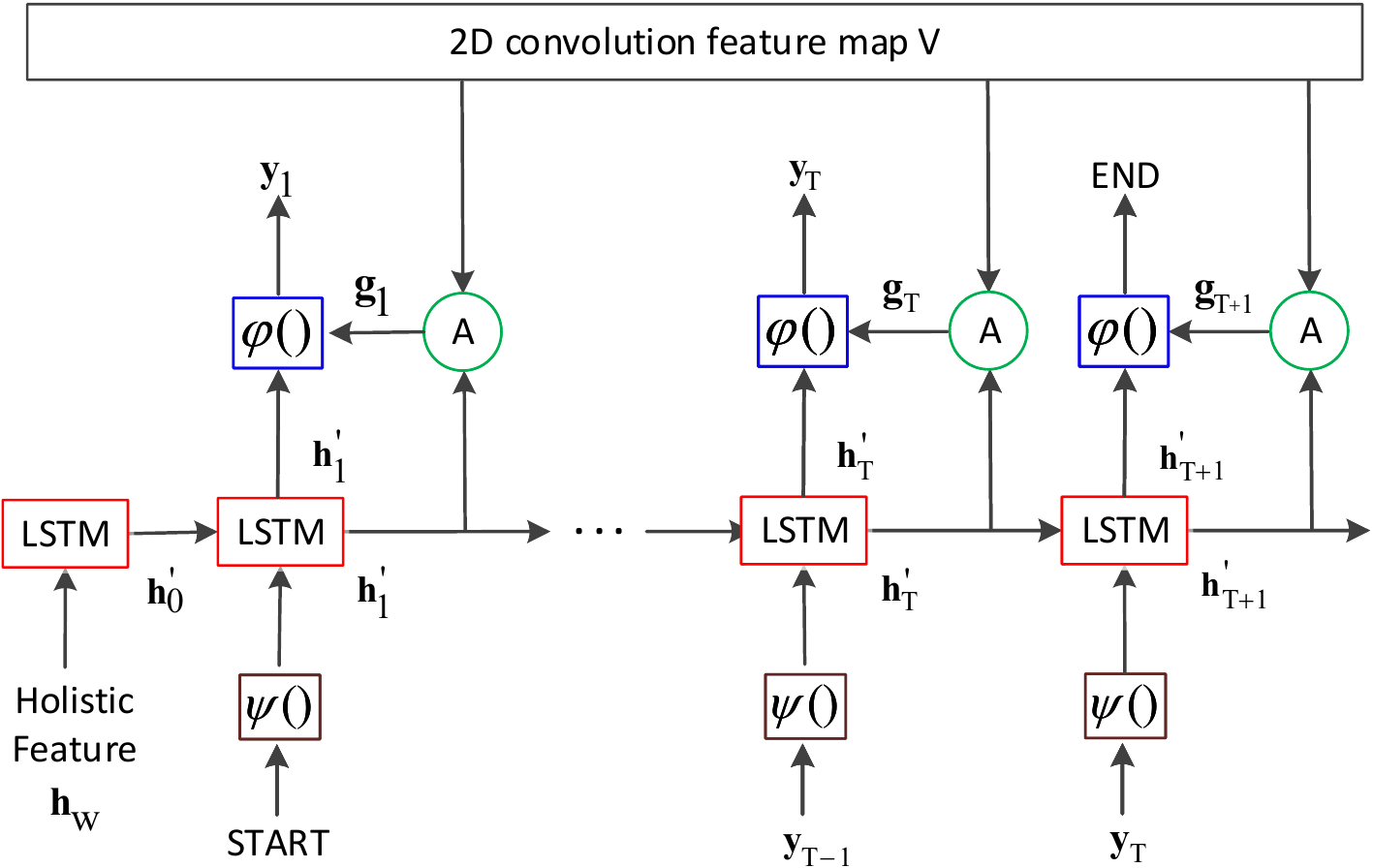}
	\end{center}
	\caption{ The structure of the LSTM decoder used in this work. The holistic feature $\mathbf{h}_{W}$, a ``START'' token and the previous outputs are input into LSTM subsequently,
		terminated by an ``END'' token. At each time step $t$, the output $y_t$ is computed by $\varphi()$ with the current hidden state and the attention output as inputs.
	}
	\label{fig:decoder}
\end{figure}

\noindent{\bf Decoder}
As shown in Figure~\ref{fig:decoder},
the decoder is another LSTM model with $2$ layers and  $512$ hidden state size per layer.
The encoder and decoder do not share parameters.
Initially, the holistic feature $\mathbf{h}_{W}$ is fed into the decoder LSTM, at time step $0$.
Then a ``START'' token is input into LSTM at step $1$.
From step $2$,
the output of the previous step is fed into LSTM until the ``END'' token is received.
All the LSTM inputs are represented by one-hot vectors,
followed by a linear transformation $\Psi()$.
During training, the inputs of decoder LSTMs are replaced by the ground-truth character sequence.
The outputs are computed by the following transformation:
\begin{equation}
\mathbf{y}_t = \mathrm{ \varphi( \mathbf{h}'_t, \mathbf{g}_t   )} = \mathrm{softmax}(\mathbf{W}_o  [ \mathbf{h}'_t; \mathbf{g}_t ] )
\end{equation}
where $\mathbf{h}'_t$ is the current hidden state and $\mathbf{g}_t$ is the output of the attention module.
$\mathbf{W}_o $ is a linear transformation, which embeds features into the output space of $94$ classes, in corresponding to $10$ digits, $52$ case sensitive letters, $31$ punctuation characters, and an ``END'' token.

\begin{figure}[t]
	\begin{center}
		\includegraphics[width=0.84\columnwidth]{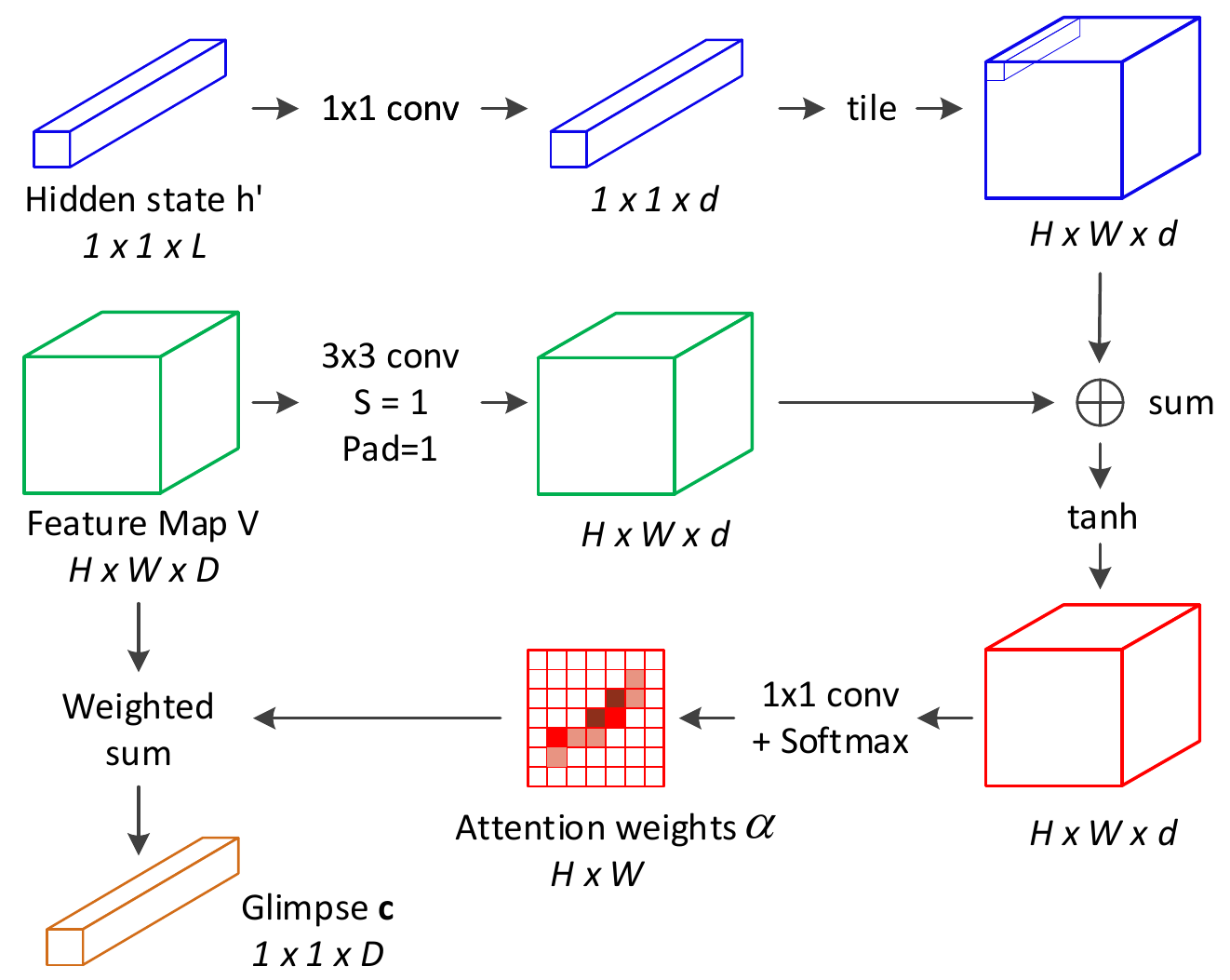}
	\end{center}
	\caption{The computation of the proposed $2$D attention mechanism can be simply implemented by convolutions, where
		$\mathbf{W}_v \mathbf{v}_{ij} + {\sum_{p,q \in \mathcal{N}_{ij}}}\! {\mathbf{\tilde{W}}_{p-i,q-j} \cdot \mathbf{v}_{pq}}$
		is accomplished by a $3\times3$ convolution. The sizes of intermediate results are also demonstrated.
		The operation ``tile'' duplicates the input $1\times1\times d$ vector $H \times W$ times.
	}
	\label{fig:attention}
\end{figure}

\noindent{\bf $2$D Attention}
Traditional $2$D attention modules~\cite{xu2015show} treat each location independently, neglecting their $2$D spatial relationships.
In order to take neighborhood information into account, we propose a tailored $2$D attention mechanism as follows:
\begin{small}
\begin{equation}
\begin{cases}
\mathbf{e}_{ij} = \tanh( \mathbf{W}_v \mathbf{v}_{ij} + \!\displaystyle{\sum_{p,q \in \mathcal{N}_{ij}}}\! {\mathbf{\tilde{W}}_{p-i,q-j} \cdot \mathbf{v}_{pq}} + \mathbf{W}_h \mathbf{h}'_t), \,\, \\
{\alpha}_{ij} = \mathrm{softmax} (\mathbf{w}_e^T \cdot \mathbf{e}_{ij}), \\
\mathbf{g}_t = \!\displaystyle{\sum_{i,j}} \, \alpha_{ij} \mathbf{v}_{ij},  \quad  i = 1, \dots, H, \quad j = 1, \dots, W.
\end{cases}
\label{eq:atten}
\end{equation}
\end{small}
where $\mathbf{v}_{ij}$ is the local feature vector at position $(i,j)$ in $\mathbf{V}$,
and $\mathcal{N}_{ij}$ is the eight-neighborhood around this position;
$\mathbf{h}'_t$ is the hidden state of decoder LSTMs at time step $t$, to be used as the guidance signal;
$\mathbf{W}_v$, $\mathbf{W}_h$
and $\mathbf{\tilde{W}}$s
are linear transformations to be learned; ${\alpha}_{ij}$ is the attention weight at location $(i,j)$; and $\mathbf{g}_t$ is the weighted sum of local features, denoted as a \textit{glimpse}.
Compared to traditional attention mechanisms, we add a term ${\sum_{p,q \in \mathcal{N}_{ij}}}\! {\mathbf{\tilde{W}}_{p-i,q-j} \cdot \mathbf{v}_{pq}}$ when computing the weight of $\mathbf{v}_{ij}$.
We can see from Figure~\ref{fig:attention} that the computation of \eqref{eq:atten} can be accomplished by a series of convolution operations.
Hence it is easy to implement.

\section{Experiments}
\label{SEC:Exp}
In this section, we perform extensive experiments to verify the effectiveness of the proposed method. We first show the datasets used for training and test,
and then demonstrate the implementation details.
Our model is compared with state-of-the-art methods on a number of public benchmark datasets, including both regular and irregular text in natural scene images.
We also conduct ablation studies to analyze the impact of model hyper-parameters and training data on performance.

\subsection{Datasets}
The following datasets are used in our experiments:

\noindent{\bf Synthetic Datasets}
There are two public available synthetic datasets that are widely used to train text recognizers: the $9$-million synthetic data (refer to as \textbf{Syn90k}) released by \cite{Max2016IJCV} and the $8$-million synthetic words (refer to as \textbf{SynthText}) proposed by \cite{Gupta16}.  Images in \textbf{Syn90k} are generated based on $90$k generic English words, while word instances in \textbf{SynthText} are from the Newsgroup$20$ lexicon~\cite{Lang95}. Although they cover a huge number of word instances, the proportion of special characters like punctuations is relatively small. To compensate the lack of special characters, we synthesize additional $1.6$-million word images (denoted as \textbf{SynthAdd}) using the synthetic engine proposed by \cite{Gupta16}. Special characters are randomly inserted to the words in the aforementioned two lexicons.

\noindent{\bf IIIT 5K-Words} (IIIT5K) ~\cite{MishraBMVC12}
contains $5000$ word patches cropped from natural scene images found by Google image search, $2000$ for training and $3000$ for test.
Text instances in these images are nearly horizontal. Each image associates with a $50$-word lexicon and a $1000$-word lexicon individually.

\noindent{\bf Street View Text} (SVT) ~\cite{Wangkai2011}
consists of $647$ word patches cropped from Google Street View for test. They are nearly horizontal, but with noise, blur and low-resolution. Each image is associated with a $50$-word lexicon.

\noindent{\bf ICDAR 2013} (IC13)~\cite{icdar2013}
has $848$ cropped word patches for training and $1095$ for test. To fairly compare with previous results,
we remove images that contain non-alphanumeric characters, which results in $1015$ test patches.
Words in this dataset are also mostly regular. No lexicon is provided.

\noindent{\bf ICDAR 2015} (IC15)~\cite{icdar2015}
contains word patches cropped from incidental scene images captured under arbitrary angles.
Hence most word patches in this dataset are irregular (oriented, perspective or curved).
It contains $4468$ patches for training and $2077$ for test. No lexicon is associated.

\noindent{\bf Street View Text Perspective} (SVTP)~\cite{SVTP}
consists of $639$ word patches, which are cropped from side-view snapshots in Google Street View and encounter severe perspective distortions.
All patches are used for test, with a $50$-word lexicon and a Full lexicon for each image.

\noindent{\bf CUTE80} (CT80)~\cite{CT80}
contains $288$ curved text images for test, with high resolution. No lexicon is associated.

\noindent{\bf COCO-Text} (COCO-T)~\cite{cocotext}
contains more than $62$k legible word patches cropped from COCO images, including machine printed and handwritten, regular and irregular text.
There are $42618$ patches for training, $9896$ for validation and $9837$ for test. No lexicon is provided.

\subsection{Implementation Details}
The proposed model is implemented in Torch. All experiments are conducted on an NVIDIA Titan X GPU with $12$GB memory.
We simply use the cross-entropy loss for training. Without any pre-training, the whole network is end-to-end trained using the ADAM optimizer~\cite{adam14}.
We use a batch size of $32$ at training time. The learning rate is set to $10^{-3}$ initially, with a decay rate of $0.9$ every $10000$ iterations until it reaches $10^{-5}$.

Iteratively, we construct distinct data groups with $120$k patches randomly sampled from \textbf{Syn90k}, $120$k from \textbf{SynthText}, $80$k from \textbf{SynthAdd}
and approximately $50$k training data from all the aforementioned public real datasets.
Each group is trained for $2$ epochs, and our algorithm converges after using $20$ groups.
In total, $2.4$ million patches from \textbf{Syn90k}, $2.4$ million from \textbf{SynthText} and $1.6$ million from \textbf{SynthAdd} are used in the whole training process.
The height of input images is resized to $48$ pixels, and the width is calculated according to the original aspect ratio, but no larger than $160$ and no smaller than $48$ pixels.

At test time, for images with height larger than width, we will rotate the image by $90$ degrees clockwise and anticlockwise respectively, and recognize them together with the original image.
A recognition score will be calculated by averaging the output probabilities. The top-scored one will be chosen as the final recognition result.
We use beam search for LSTM decoding, which keeps the top-$k$ candidates with the highest accumulative scores, where $k$ is empirically set to $5$ in our experiments.
Compared to greedy decoding that only picks the highest scored character at each time step,
beam search brings an approximately $0.5\%$ improvement to the recognition accuracy, in our practice.
The test speed is $15$ms per patch in average.

\begin{table*}[ht!]
	\newcommand{\tabincell}[2]{\begin{tabular}{@{}#1@{}}#2\end{tabular}}
	\begin{center}
		\caption{Recognition accuracy (in percentages) on public benchmarks, including both regular and irregular text. ``50'', ``1k'', and ``Full'' are lexicon sizes, where ``Full'' means a combined lexicon of all images in the dataset. ``None'' means lexicon-free. The approaches marked with ``*'' are trained with both word-level and character-level annotations. In each column, the best performing result is shown in $\textbf{bold}$ font, and the second best result is shown with underline. Our approach outperforms all the compared methods on all irregular text benchmarks, and achieves comparable performance on regular text. }
		\label{Tab:1}
		\resizebox{1.87\columnwidth}{!}{
			\begin{tabular}{l|c|c|c|c|c|c|c|c|c|c|c|c}
				\hline
				Method & \multicolumn{6}{|c}{Regular Text} &  \multicolumn{6}{|c} {Irregular Text}   \\
				\cline{2-13} & \multicolumn{3}{c|}{IIIT5K} & \multicolumn{2}{|c|}{SVT}  & \multicolumn{1}{|c|}{IC13}   & \multicolumn{1}{c|}{IC15} & \multicolumn{3}{|c|}{SVTP} & \multicolumn{1}{|c|}{CT80} & \multicolumn{1}{|c}{COCO-T}   \\
				\cline{2-13} & \multicolumn{1}{|c|}{50}  & \multicolumn{1}{|c|}{1k}   & \multicolumn{1}{c|}{None}  & \multicolumn{1}{|c|}{50}  & \multicolumn{1}{|c|}{None} & \multicolumn{1}{|c|}{None} & \multicolumn{1}{|c|}{None} & \multicolumn{1}{|c|}{50} & \multicolumn{1}{|c|}{Full} & \multicolumn{1}{|c|}{None} &\multicolumn{1}{|c}{None} & \multicolumn{1}{|c}{None} \\
				\hline
				\cite{Wangkai2011} & $-$ & $-$ & $-$  & $57.0$  & $-$ & $-$ & $-$ & $40.5$ & $21.6$ & $-$ & $-$ & $-$   \\
				\hline
				\cite{MishraCVPR2012} & $64.1$ & $57.5$ & $-$  & $73.2$  & $-$ & $-$ & $-$ & $45.7$ & $24.7$ & $-$ & $-$ & $-$ \\
				\hline
				\cite{SVTP} & $-$ & $-$ & $-$  & $73.7$  & $-$ & $-$ & $-$ & $75.6$ & $67.0$ & $-$ & $-$  \\
				\hline
				\cite{YaoCVPR2014} & $80.2$ & $69.3$ & $-$  & $75.9$  & $-$ & $-$ & $-$ & $-$ & $-$ & $-$ & $-$  \\
				\hline
				\cite{Max2016IJCV} & $97.1$ & $92.7$ & $-$  & $95.4$  & $80.7$ & $90.8$ & $-$ & $-$ & $-$ & $-$ & $42.7$  \\
				\hline
				\cite{He2015Reading} & $94.0$ & $91.5$ & $-$  & $93.5$  & $-$ & $-$ & $-$ & $-$ & $-$ & $-$ & $-$  \\
				\hline
				\cite{Lee_2016_CVPR} & $96.8$ & $94.4$ & $78.4$  & $96.3$  & $80.7$ & $90.0$ & $-$ & $-$ & $-$ & $-$ & $-$  \\
				\hline
				\cite{OCRNIPS17} & $98.0$ & $95.6$ & $80.8$  & $96.3$  & $81.5$ & $-$ & $-$ & $-$ & $-$ & $-$ & $-$  \\
				\hline
				\cite{shiCVPR2016} & $96.2$ & $93.8$ & $81.9$  & $95.5$  & $81.9$ & $88.6$ & $-$ & $91.2$ & $77.4$ & $71.8$ & $59.2$ & $-$   \\
				\hline
				\cite{BMVC2016_43} & $97.7$ & $94.5$ & $83.3$  & $95.5$  & $83.6$ & $89.1$ & $-$ & $\underline{94.3}$ & $83.6$ & $73.5$ & $-$  \\
				\hline
				\cite{ShiBY15} & $97.8$ & $95.0$ & $81.2$  & $97.5$  & $82.7$ & $89.6$ & $-$ & $92.6$ & $72.6$ & $66.8$ & $54.9$ & $-$   \\
				\hline
				\cite{ijcai2017}* & $97.8$ & $96.1$ & $-$  & $95.2$  & $-$ & $-$ & $-$ & $93.0$ & $80.2$ & $75.8$ & $69.3$ & $-$   \\
				\hline
				\cite{Cheng2017}* & $99.3$ & $97.5$ & $87.4$  & $97.1$  & $85.9$ & $93.3$ & $70.6$ & $92.6$ & $81.6$ & $71.5$ & $63.9$ & $-$   \\
				\hline
				\cite{SqueezeText18}* & $97.0$ & $94.1$ & $87.0$  & $95.2$  & $-$ & $92.9$ & $-$ & $-$ & $-$ & $-$ & $-$  \\
				\hline
				\cite{Liu2018CharNetAC}* & $-$ & $-$ & $92.0$  & $-$  & $85.5$ & $91.1$ & $74.2$ & $-$ & $-$ & $\underline{78.9}$ & $-$ & $\underline{59.3}$  \\
				\hline
				\cite{cheng_EditDistance}* & $\underline{99.5}$ & $97.9$ & $88.3$  & $96.6$  & $87.5$ & $\textbf{94.4}$ & $73.9$ & $-$ & $-$ & $-$ & $-$  \\
				\hline
				\cite{Cheng2018AON} & $\textbf{99.6}$ & $98.1$ & $87.0$  & $96.0$  & $82.8$ & $-$ & $68.2$ & $94.0$ & $\underline{83.7}$ & $73.0$ & $76.8$  \\
				\hline
				\cite{shiPAMI2018} & $\textbf{99.6}$ & $\textbf{98.8}$ & $\underline{93.4}$  & $\textbf{99.2}$  & $\textbf{93.6}$ & $91.8$ & $\underline{76.1}$ & $-$ & $-$ & $78.5$ & $\underline{79.5}$  \\
				\hline
				\hline
				SAR (Ours) & $99.4$ & $\underline{98.2}$ & $\textbf{95.0}$  & $\underline{98.5}$  & $\underline{91.2}$ & $\underline{94.0}$  & $\textbf{78.8}$  & $\textbf{95.8}$ & $\textbf{91.2}$ & $\textbf{86.4}$  & $\textbf{89.6}$ & $\textbf{66.8}$  \\
				\hline
			\end{tabular}
		}
	\end{center}
\end{table*}

\begin{figure*}[th!]
	\begin{center}
		\includegraphics[width=0.67009\textwidth]{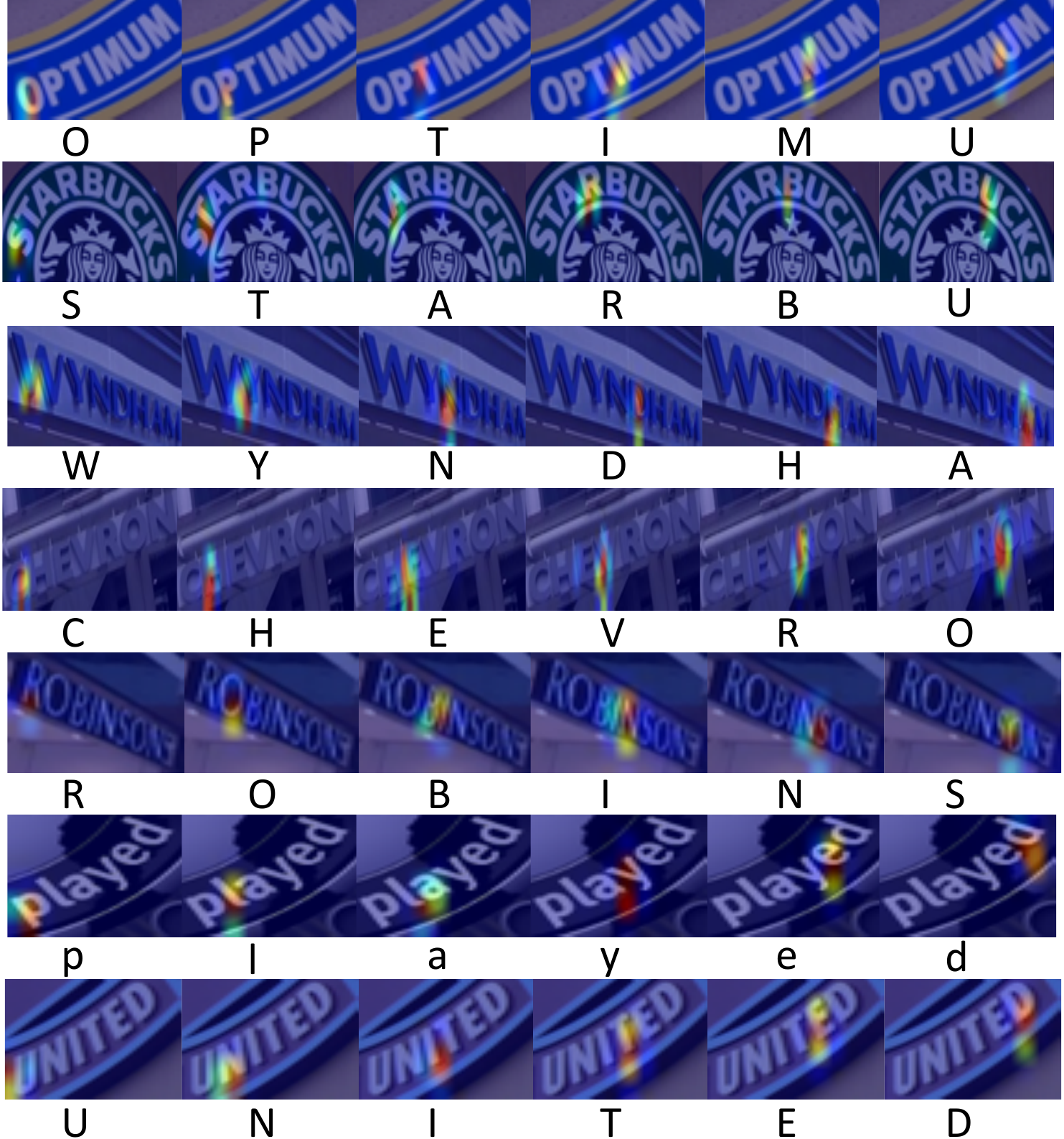}
	\end{center}
	\caption{
		Visualization of $2$D attention weights at individual decoding time steps, which shows that our $2$D attention model can be trained to approximately localize characters without character-level annotations. For space reasons, some of the decoding results are truncated.
	}
	\label{fig:visualization}

\end{figure*}

\begin{figure*}[t!]
	\begin{center}
		\includegraphics[width=0.67009\textwidth]{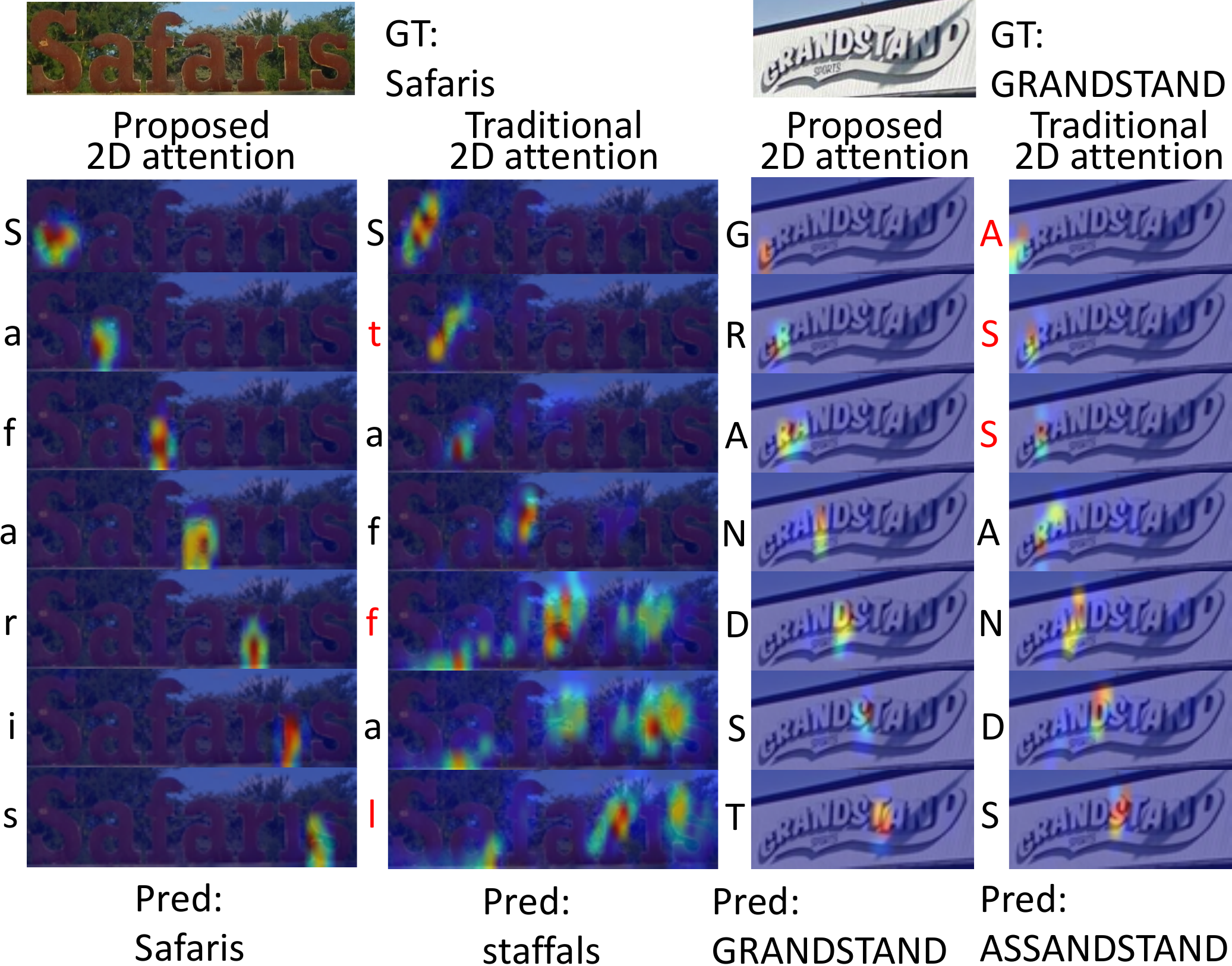}
	\end{center}
	\caption{Comparison of our proposed $2$D attention model and the traditional $2$D attention model.
		The decoded characters are shown to the left of the corresponding attention heat maps, with incorrect ones marked in red.
		The proposed model shows more accurate localization and better recognition results.
	}
	\label{fig:2atten}
\end{figure*}

\begin{figure}[t!]
	\begin{center}
		\includegraphics[width=0.85\columnwidth]{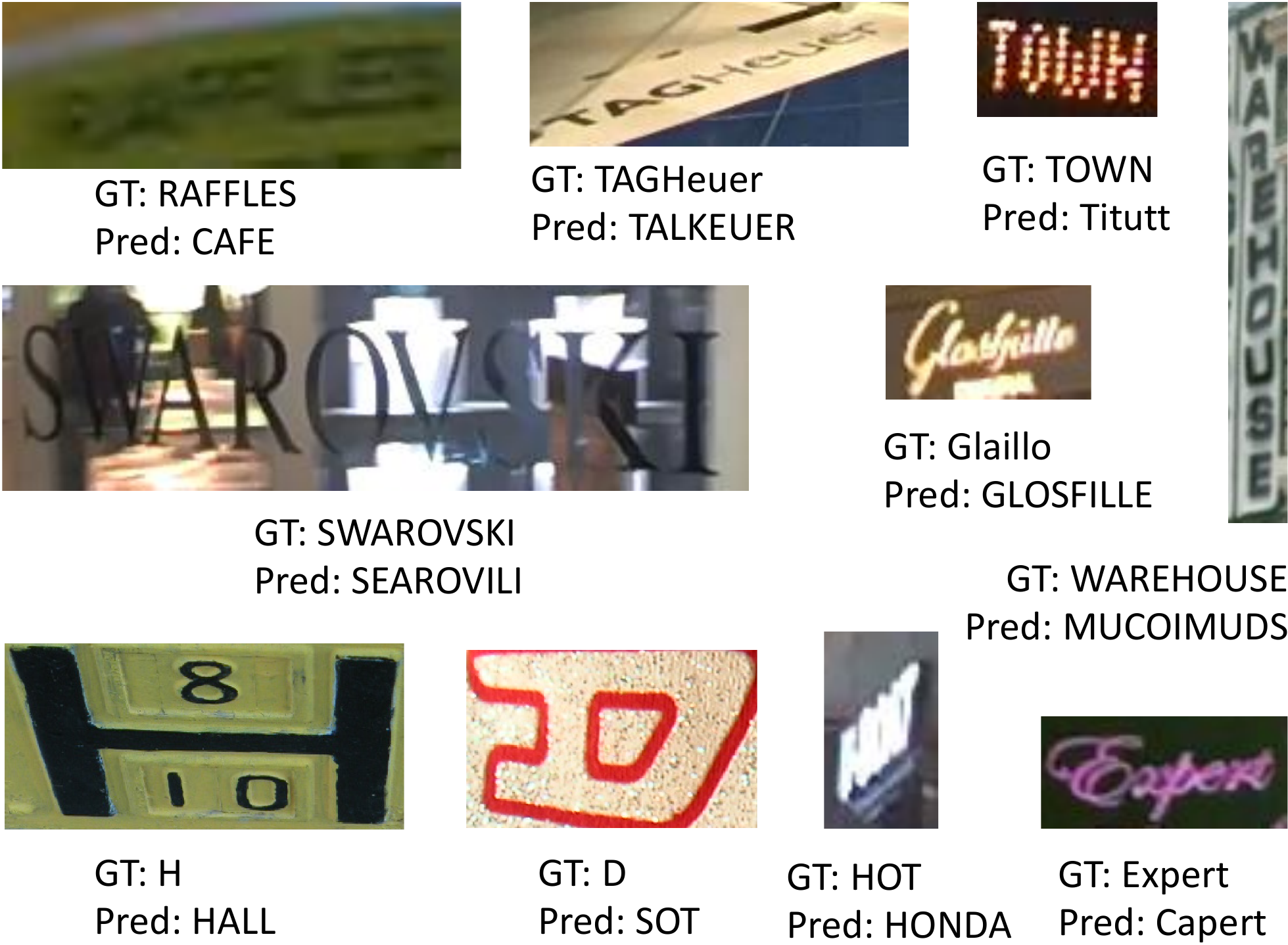}
	\end{center}
	\caption{Failure cases of our model. ``GT'' stands for the ground-truth annotation, and ``Pred'' is the predicted result.
	}
	\label{fig:failure}
\end{figure}

\subsection{Experimental Results}

\begin{table*}[th!]
	\newcommand{\tabincell}[2]{\begin{tabular}{@{}#1@{}}#2\end{tabular}}
	\begin{center}
		\caption{Ablation studies by changing model hyper-parameters and training data. The first row shows the original proposed model configuration.
		Those changed parameters are shown in {\bf bold} font. The models are evaluated on benchmarks without using any lexicon.
	    Reducing the size of CNN and LSTM models has negative impacts on the recognition performance. Using the traditional $2D$ attention or $1$D attention
	    modules, instead of our proposed attention mechanism, also degrades the accuracy.
        }
		\label{Tab:2}
			\resizebox{1.9\columnwidth}{!}{
		\begin{tabular}{c|c|c|c|c|c|c|c|c|c|c|c|c}
			\hline
			\multicolumn{6}{c|}{Model Configuration}  &&&&&&&  \\
			\cline{1-6} \multicolumn{1}{c|}{Training } & \multicolumn{1}{c|}{CNN} &\multicolumn{1}{c|}{Down-sampling}  & \multicolumn{1}{|c|}{Attention}   & \multicolumn{1}{|c|}{LSTM}  & \multicolumn{1}{|c|}{Hidden state}  &  IIIT5K & SVT  & IC13  & IC15 & SVTP & CT80 & COCO-T
			\\
			\multicolumn{1}{c|}{data} &\multicolumn{1}{c|}{channels} &\multicolumn{1}{c|}{ratio}  & \multicolumn{1}{|c|}{module}   & \multicolumn{1}{|c|}{layers}  & \multicolumn{1}{|c|}{size}  &&&&&&&
			\\
			\hline
			\multirow{9}*{Synth+Real} & $\times 1$ &$1/8$, $1/4$ & $2$D proposed & $2$ & $512$ & $95.0$ & $91.2$ & $94.0$ & $78.8$ &$86.4$ & $89.6$ & $66.8$ \\
			\cline{2-13}
		    & $\times \mathbf{{1}/{2}}$ &$1/8$, $1/4$ & $2$D proposed & $2$ & $512$ & $92.7$ & $88.7$ & $92.0$ & $75.6$ & $81.3$ & $86.8$&  $62.6$\\
			& $\times 1$ &$\mathbf{1/16}$, $1/4$ & $2$D proposed & $2$ & $512$ &$93.8$ & $90.3$ & $92.7$ &$77.4$ & $84.5$ & $89.2$ & $64.8$ \\
			 & $\times 1$ &$\mathbf{1/16}$, $\mathbf{1/8}$ & $2$D proposed & $2$ & $512$ & $94.0$ & $90.6$ & $93.1$ & $76.2$ & $83.7$ & $87.5$ & $63.7$\\
			& $\times 1$ &$1/8$, $\mathbf{1/8}$ & $2$D proposed & $2$ & $512$  &$93.6$ & $89.3$ & $92.5$ & $76.1$ & $82.8$ & $87.5$ & $63.3$\\
				\cline{2-13}
			& $\times 1$ &$1/8$, $1/4$ & \bf{$\mathbf{2}$D traditional} & $2$ & $512$ & $94.0$ & $90.1$ & $92.3$ & $77.2$ & $84.3$ & $87.5$ &$64.2$\\
		   & $\times 1$ &$1/8$, $1/4$ & \bf{$\mathbf{1}$D}  & $2$ & $512$ & $93.0$ & $89.9$ &  $90.2$ & $76.6$ & $83.6$ & $84.7$ & $65.4$\\
				\cline{2-13}
			 & $\times 1$ &$1/8$, $1/4$ & $2$D proposed & $\mathbf{1}$ & $512$ & $89.7$ & $87.2$ & $87.4$ & $70.6$ & $76.4$ & $80.6$&  $60.1$\\
			 & $\times 1$ &$1/8$, $1/4$ & $2$D proposed & $2$ & $\mathbf{256}$ & $94.0$ & $89.3$ & $92.8$ & $76.8$ & $83.7$ & $86.5$ & $63.8$ \\
			\hline
			\textbf{OnlySynth} & $\times 1$ &$1/8$, $1/4$ & $2$D proposed & $2$ & $512$ & $91.5$ & $84.5$ & $91.0$ & $69.2$ &$76.4$ & $83.3$ & $-$ \\
			\hline
		\end{tabular}
		}

	\end{center}
\end{table*}

In this section, we evaluate our model on several regular and irregular text benchmarks, and compare the performance with other state-of-the-art methods.
For datasets with lexicons provided, we simply select from lexicon the one with the minimum edit distance to the predicted word.
The recognition results are summarized in Table~\ref{Tab:1}.

On irregular text datasets (\ie, IC15, SVTP, CT80 and COCO-T),
our approach outperforms the compared methods by a large margin.
In particular, our approach gives accuracy increases of $7.5\%$ ($78.9\%$ to $86.4\%$) on SVTP-None and $10.1\%$ ($79.5\%$ to $89.6\%$) on CT80.
Note that neither SVTP or CT80 provides training data, which lowers the chance of over-fitting.
Meanwhile, the proposed method still achieves state-of-the-art performance on regular text datasets (\ie, IIIT5K, SVT and IC13).
Actually, our model performs the best or the second best on $5$ of the $6$ evaluated regular text settings.
The superiority of our method is more significant when there is no lexicon, such as in IIIT5K and SVTP.
It demonstrates the practicality of our proposed approach in realistic scenarios where lexicon is rarely provided.

Examples of $2$D attention heat maps when decoding individual characters are visualized in Figure~\ref{fig:visualization}.
Although learned in a weakly supervised manner, the attention module can still approximately localize characters being decoded,
extract discriminative local features and finally help text recognition. Note that the proposed attention module is trained without character-level annotations.

Some failure cases are also presented in Figure~\ref{fig:failure}.
There are a variety of reasons for failure, such as blurry, partial occlusion, extreme distortion, uneven lighting condition, uncommon fonts, vertical text, \etc.
Scene text recognition still has a long way to be completely solved.

\subsection{Ablation Studies}

In order to analyze the impact of different model hyper-parameters and training data on the recognition performance, we perform a series of ablation studies as presented in Table~\ref{Tab:2}.
All the evaluated models are trained from scratch and tested on benchmarks without lexicon.

\subsubsection{\bf CNN Parameters}
We firstly reduce by $50\%$ the number of channels of all convolutional layers expect the last layer, which lowers the accuracy by $2$ to $4$ percentages.
The down-sampling ratio of the proposed ResNet is $1/8$ vertically and $1/4$ horizontally, which results in a feature map of maximum size $6 \times 40$.
Here we further divide the vertical and/or horizontal down-sampling ratios by $2$, and obtain worse performance.
These results shows that the volume of feature maps should be sufficiently large to encode a large variety of visual information for text recognition.

\subsubsection{\bf Attention Modules}
The proposed $2$D attention model is respectively replaced by the traditional $2$D counterpart with the term ${\sum_{p,q \in \mathcal{N}_{ij}}}\! {\mathbf{\tilde{W}}_{p-i,q-j} \cdot \mathbf{v}_{pq}}$ removed from Equation~\eqref{eq:atten} and a $1$D attention module that considers feature maps as $1$D sequences.
By aggregating neighborhood information, the proposed $2$D attention model outperforms the traditional $2$D one by $1$ to $2$ percentages.
Both of the $2$D attention modules performs better than the $1$D one in most cases, which shows their robustness for both regular and irregular text recognition.
We compare the proposed and the traditional $2$D attention heat maps in Figure~\ref{fig:2atten}.
The proposed model presents better performance on character localization and recognition.

\subsubsection{\bf LSTM Parameters}
By cutting down by half the hidden state size of both encoder and decoder LSTMs, we receive degraded recognition accuracies. The performance degradation is more serious when we use $1$ layer of LSTMs, instead of $2$ layers.
Relatively, the number of LSTM layers presents a stronger impact on the performance.

\subsubsection{\bf Training Data}
The performance drops without using the real image training data. However, because of our simple model design, we can make full use of current public available dataset to train the model for better performance. Moreover, it should be note that a large number of methods (as indicated with ``*'' in Table~\ref{Tab:1}) use extra character level annotations, which is another kind of additional information. The requirement of character-level annotations prevents them from using real data with only word-level annotations.

\section{Conclusion}
\label{SEC:Con}
In this work, we present a simple yet strong baseline for irregular text recognition.
The proposed framework is built upon off-the-shelf neural network modules, including a ResNet CNN, an LSTM encoder-decoder and
a $2$D tailored attention module.
Without any extra supervision information, the proposed attention mechanism is capable of select local features for decoding characters.
Being robust to different forms of text layouts, our approach performs well for both regular and irregular text.

As to future works, the proposed framework can be extended in several ways.
Firstly, the LSTM encoder-decoder is possible to be replaced by CNNs for sequence modeling, which will further ease the training process.
Secondly, the proposed $2$D attention module can be seen as a special case of graph neural networks, where edges of the graph are defined on $8$-neighborhoods.
Straightforwardly, we can apply the attention mechanism on graphs with more complex structures, to incorporate with the rich context information.
Finally, to better learn visual features and speedup the training process, we can also add a word classification head apart from the LSTM decoder.

\bibliographystyle{aaai}
\bibliography{MyBibFile}

\end{document}